\documentclass{article}

\usepackage{microtype}
\usepackage{graphicx}
\usepackage{subcaption}
\usepackage{booktabs} %

\usepackage{hyperref}
\usepackage{url}
\usepackage{amssymb}
\usepackage{amsmath}
\usepackage{multirow}
\usepackage{enumitem}
\usepackage{tablefootnote}
\usepackage{makecell}
\usepackage{comment}
\usepackage{booktabs}
\usepackage{bbm}
\usepackage{wrapfig}
\usepackage{graphicx}
\usepackage[most]{tcolorbox}   %
\usepackage[table]{xcolor}  %
\usepackage{soul}              %
\usepackage{marginnote}        %
\usepackage{textcomp}
\usepackage{minted}
\usepackage{changepage}
\usepackage{listings}
\usepackage[T1]{fontenc}

\newcommand{\hi}[1]{\textcolor{blue}{\textbf{#1}}}

\usepackage[preprint]{icml26/icml2026}

\usepackage{amsmath}
\usepackage{amssymb}
\usepackage{mathtools}
\usepackage{amsthm}

\usepackage[capitalize,noabbrev]{cleveref}

\theoremstyle{plain}

\theoremstyle{definition}

\theoremstyle{remark}

\usepackage[textsize=tiny]{todonotes}

\icmltitlerunning{ContextEvolve: Multi-Agent Context Compression for Systems Code Optimization}

\begin{document}

\twocolumn[
    \icmltitle{ContextEvolve: Multi-Agent Context Compression for Systems Code Optimization}

  \icmlsetsymbol{equal}{*}

  \begin{icmlauthorlist}
    \icmlauthor{Hongyuan Su}{thu,zgc}
    \icmlauthor{Yu Zheng}{mit}
    \icmlauthor{Yong Li}{thu,zgc}
  \end{icmlauthorlist}

\icmlaffiliation{thu}{Tsinghua University, Beijing, China}
\icmlaffiliation{zgc}{Zhongguancun Academy, Beijing, China}
\icmlaffiliation{mit}{Massachusetts Institute of Technology, Cambridge MA, USA}

  \vskip 0.3in
]

\printAffiliationsAndNotice{}  %

\begin{abstract}

Large language models are transforming systems research by automating the discovery of performance-critical algorithms for computer systems.
Despite plausible codes generated by LLMs, producing solutions that meet the stringent correctness and performance requirements of systems demands iterative optimization.
Test-time reinforcement learning offers high search efficiency but requires parameter updates infeasible under API-only access, while existing training-free evolutionary methods suffer from inefficient context utilization and undirected search.
We introduce \textbf{ContextEvolve}, a multi-agent framework that achieves RL-level search efficiency under strict parameter-blind constraints by decomposing optimization context into three orthogonal dimensions: a \textit{Summarizer Agent} condenses semantic state via code-to-language abstraction, a \textit{Navigator Agent} distills optimization direction from trajectory analysis, and a \textit{Sampler Agent} curates experience distribution through prioritized exemplar retrieval.
This orchestration forms a functional isomorphism with RL—mapping to state representation, policy gradient, and experience replay—enabling principled optimization in a textual latent space.
On the ADRS benchmark, ContextEvolve outperforms state-of-the-art baselines by \textbf{33.3\%} while reducing token consumption by \textbf{29.0\%}.
Codes for our work are released at \href{https://anonymous.4open.science/r/ContextEvolve-ACC}{\textcolor{blue}{https://anonymous.4open.science/r/ContextEvolve-ACC}}.

\end{abstract}

\section{Introduction}

Systems research has traditionally relied on human experts to design algorithms that optimize performance under strict constraints~\cite{zoph2016neural,jiang2024llmopt}.
The emergence of Large Language Models (LLMs) has enabled \textit{AI-Driven Research for Systems} (ADRS)~\cite{cheng2025barbarians,jiang2024survey}, where LLMs automate the design of solutions for databases, networking, and distributed systems~\cite{liu2025supporting,yu2025prism,wooders2024cloudcast}.
While LLMs can generate plausible code candidates, the stringent correctness and performance requirements of systems demand \textit{rigorous iterative refinement}.
Consequently, research focus has shifted from one-shot generation toward autonomous multi-round optimization~\cite{jiang2024self}: iteratively improving solutions until they exceed human-engineered baselines or exhaust computational budgets.

Test-time reinforcement learning (RL) offers a natural framework for such optimization via an RL loop of solution generation, reward evaluation, and LLM parameter update~\cite{hubert2025olympiad}.
However, despite its promising performance, the scale of modern LLMs makes parameter updates computationally prohibitive~\cite{kaplan2020scaling}.
Meanwhile, leading LLM providers typically offer exclusively API-based access, precluding weight entirely.
Training-free alternatives—evolutionary strategies such as AlphaEvolve~\cite{novikov2025alphaevolve} and multi-agent frameworks such as CAMEL~\cite{li2023camel}—sidestep this constraint but exhibit \textit{low search efficiency}: they lack mechanisms for compressing accumulated context during iteration and for extracting precise optimization signals from long evolutionary history.

In this paper, we introduce \textbf{ContextEvolve}, a multi-agent framework designed for high search efficiency under parameter-blind constraints.
Our key insight is that, instead of relying on a single monolithic model to manage the entire search state, optimization context can be decomposed into three orthogonal dimensions, each managed by a specialized agent.
In ContextEvolve, a \textit{Summarizer Agent} condenses semantic state, distilling code artifacts into natural language abstracts preserving critical state information; a \textit{Navigator Agent} distills optimization direction, extracting textual gradients from trajectory analysis; and a \textit{Sampler Agent} manages experience distribution, retrieving diverse, high-value exemplars as few-shot references.
Crucially, this orchestration establishes a \textit{functional isomorphism} with test-time RL: the Summarizer Agent corresponds to state representation learning, the Navigator Agent emulates policy gradient estimation, and the Sampler Agent implements prioritized experience replay.
This alignment enables ContextEvolve to inherit RL's sample efficiency while operating entirely in a text space without parameter access.

Our main contributions are:
\begin{itemize}[nosep,leftmargin=1.5em]
    \item We propose ContextEvolve, a multi-agent framework achieving high search efficiency for system code optimization under API-only constraints via \textit{structured context compression}.
    \item We introduce a suite of three specialized agents, including Summary, Gradient, and Sampler, which decompose context into semantic state, optimization direction, and experience distribution, and collectively approximate a test-time RL loop in a parameter-blind and textual space.
    \item On the ADRS benchmark across diverse systems code optimization domains, ContextEvolve surpasses state-of-the-art methods by \textbf{33.3\%} while reducing token consumption by \textbf{29.0\%}.
\end{itemize}

\section{Related Works}

\noindent\textbf{Context Management.}
Long-horizon agentic workflows quickly accumulate large search state information, making context management a central bottleneck.
Prior work addresses it from several angles: scaling model-side context capacity despite common failures such as positional bias and the lost-in-the-middle effect~\citep{dai2019transformerxl,beltagy2020longformer,zaheer2020bigbird,borgeaud2021retro,liu2023lost,zhang2024found}, externalizing long-term state into system-side memory~\citep{lewis2020rag,park2023generativeagents,packer2023memgpt}, increasing information density via prompt compression and token pruning~\citep{jiang2023llmlingua,jiang2023longllmlingua,pan2024llmlingua2}, and reducing prompt burden through agentic structuring that decomposes reasoning, search, and tool interaction into explicit loops~\citep{yao2022react,yao2023tot,shinn2023reflexion,yang2024sweagent}.
However, these general approaches do not explicitly separate the distinct requirements of evolutionary systems code optimization, including preserving functional invariants, extracting improvement direction from noisy multi-metric trajectories, and maintaining diversity.
In contrast, ContextEvolve decomposes context into semantic state, optimization direction, and experience distribution and assigns each to a specialized agent for high information density and low token cost across long runs.

\noindent\textbf{LLM-based Evolutionary.}
For high-stakes code generation, one-shot LLM outputs often exhibit incorrectness or underperform test-time search loops that repeatedly propose candidates, evaluate with automated verifier, and update future proposals from historical feedback.
Recent work instantiates this with evolutionary coding agents that treat programs as individuals and use evaluation-driven selection and mutation for open-ended evolution and algorithm discovery~\citep{lange2025shinkaevolve,codeevolve2025,wan2025loongflow,wu2024evolutionary,liu2024large}.
Closely related, prompt/instruction evolution optimizes discrete textual artifacts via population-based search and reflective mutation/selection~\citep{fernando2023promptbreeder,guo2023evoprompt,zhou2022ape,pryzant2023apo,yang2023large}; orthogonally, several approaches pursue more \emph{directed} improvement signals without weight updates by storing verbal feedback or optimizing against model-provided critiques~\citep{shinn2023reflexion,yuksekgonul2025optimizing,zhang2025codegrad}.
Nevertheless, existing evolutionary pipelines remain token-inefficient because they inflate prompts with raw history or compress context without explicit semantic disentanglement, leading to slower iteration and higher cost.
Our proposed ContextEvolve targets these bottlenecks by maintaining compact semantic state, distilling textual gradients from weighted trajectory rollouts, and sampling exemplars via an RL-inspired experience distribution mechanism.

\section{Preliminaries}

\subsection{Single-shot Generation}
Let $\mathcal{C}$ denote the potentially discrete space of valid executable code solutions for a given problem.
Given a task description $\mathcal{D}$, a trainable LLM $\mathcal{M}_\theta$ produces executable code $c \in \mathcal{C}$ via the sampling process $c \sim \mathcal{M}_\theta( \cdot \mid \mathcal{D})$. 
Then an automated evaluation oracle $\mathcal{E}: \mathcal{C} \rightarrow \mathbb{R}$ evaluates the code and returns a scalar score $s=\mathcal{E}(c)$ reflecting the solution quality.

\subsection{Evolutionary Optimization}
Training-free evolutionary methods treat the model parameters $\theta$ as immutable, and extend the single-shot paradigm to an iterative evolutionary process where the system iteratively refines solutions based on historical feedback.
Define the optimization history $\mathcal{H}_t$ at step $t$ as the sequence of generated solutions and their scores up to the previous step,
\begin{equation}
    \mathcal{H}_t = \{(c_0, s_0), (c_1, s_1), \dots, (c_{t-1}, s_{t-1})\},
\end{equation}
where each $c_i \in \mathcal{C}$ and $s_i = \mathcal{E}(c_i)$, and a context compression operator $\Phi$ that distills $\mathcal{H}_t$ into a short textual representation.
The generation process at step $t$ could be thus formulated as,
\begin{equation}
    c_t \sim \mathcal{M}_\theta(\cdot \mid \Phi(\mathcal{H}_t), \mathcal{D}), \quad t \leq T,
\end{equation}
where $T$ is the maximum iteration.

Our goal is to design the context compression operator $\Phi$ to maximize the expected score of the best solution,
\begin{equation}
    \text{Maximize} \ \mathbb{E} \left[ \max_{i=0}^{T} \mathcal{E}(c_i)  \mid \Phi,\mathcal{D} \right].
\end{equation}

\section{Method}

\begin{figure*}
    \centering
    \includegraphics[width=0.88\linewidth]{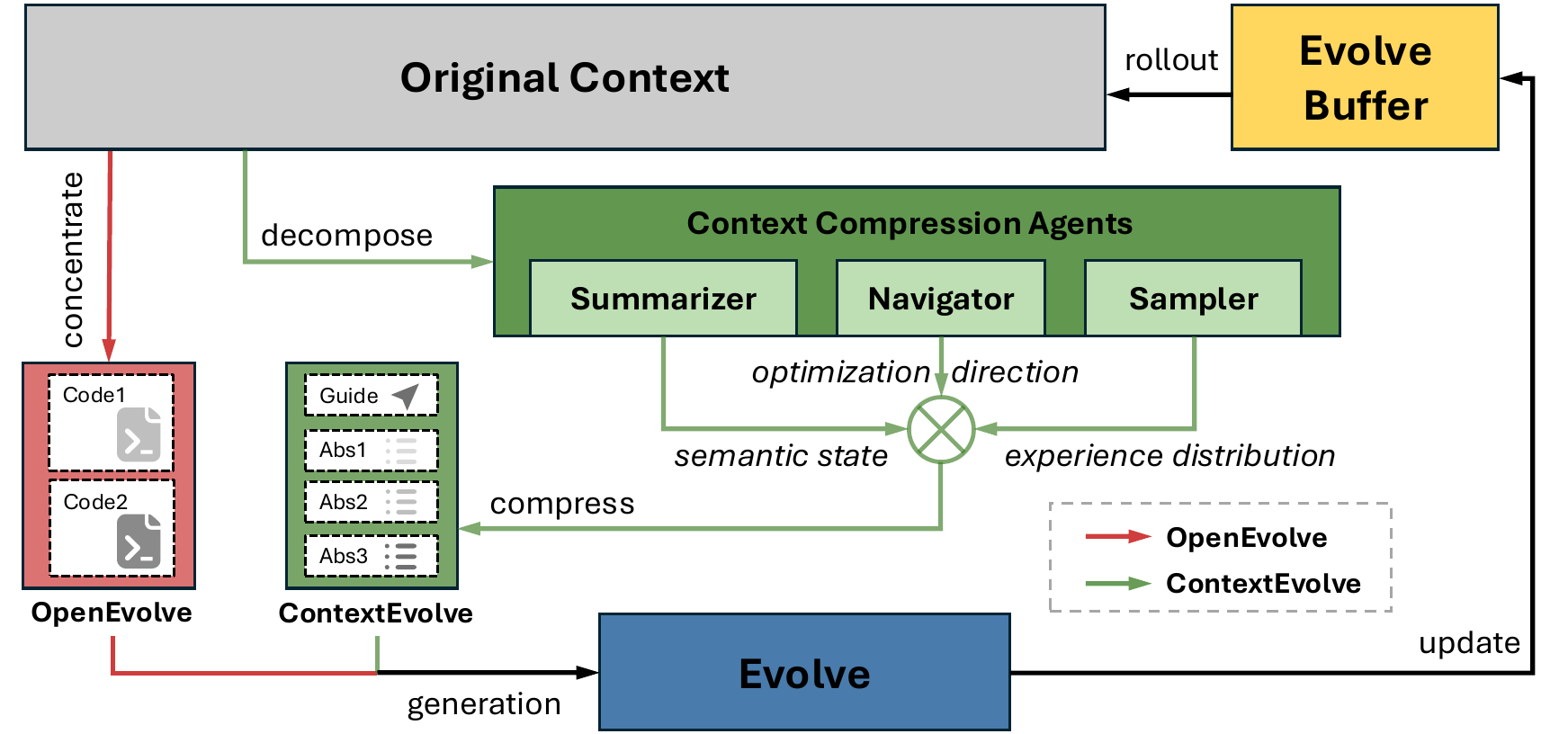}
    \caption{The pipeline comparison of ContextEvolve and OpenEvolve.
     OpenEvolve (red) directly concentrates the few original codes in limited context window, leading to low information density and inefficient evolutionary search.
     ContextEvolve (green) leverages specialized context distillation agents to compress the lengthy context, enriching the limited window with numerous valuable information.
    }
    \label{fig:pipeline}
    \vspace{-12px}
\end{figure*}

\subsection{Overall Framework}

We propose ContextEvolve, a multi-agent framework for ADRS that compresses raw interaction logs into a refined optimization context, thereby enabling high search efficiency under strict parameter-blind constraints.
To effectively compress the extensive evolutionary experience into a finite window, our framework decomposes the optimization context into three orthogonal dimensions: \textit{semantic state}, \textit{optimization direction}, and \textit{experience distribution}.
Accordingly, we introduce three specialized agents to distinctively manage them.
Specifically, the Summarizer Agent maintains the \textit{semantic state} by condensing complex code artifacts into concise natural language abstracts.
The Navigator Agent steers the \textit{optimization direction} by distilling promising search guidance from historical trajectories.
Finally, the Sampler Agent modulates the \textit{experience distribution} by curating diverse and high-value exemplars to serve as instructive few-shot references.
This decomposition and agent specialization allow our approach to construct a compact, semantically rich context representation that effectively utilizes previous experience for the next generation.

Beyond context compression, we also establish a functional isomorphism with RL algorithms, where our agents collaboratively approximate key mechanisms of search efficiency purely in the natural language space without any parameter updates.
Specifically, the Summarizer Agent corresponds to state representation learning, the Navigator Agent emulates policy gradient updates, and the Sampler Agent implements prioritized experience replay.
This theoretical alignment ensures that our framework is not merely a heuristic search but a principled isomorphism of RL, thereby significantly highlighting the search efficiency of ContextEvolve.

The collaborative workflow of these agents constitutes our evolutionary pipeline, as detailed in Algorithm~\ref{alg:contextevolve}.
In each iteration, the process begins by selecting a parent solution $c^p$ from the Evolve Buffer based on pre-defined criteria.
First, the Navigator Agent analyzes multiple evolution trajectories rolled out from the Evolve Buffer.
By scrutinizing the correlation between code modifications and metric fluctuations, it distills optimization directions for potential performance gains.
Next, conditioned on the parent's semantic state and this directional guidance, the Sampler Agent curates a set of instructive exemplars from the population to serve as few-shot references.
Subsequently, the Generator Agent integrates the parent, the directional guidance, and the retrieved exemplars to generate a superior offspring $c^c$, which is immediately assessed by the evaluator for its fitness.
Following evaluation, the Summarizer Agent compares the offspring against its parent to abstract the key characteristics into a new semantic summary.
Finally, the new generated code along with its abstract and fitness will be added to the Evolve Buffer and ready for the next iteration.

\begin{algorithm}[t]
   \caption{ContextEvolve: Multi-Agent Evolutionary Pipeline (with RL Functional Isomorphism)}
   \label{alg:contextevolve}
\begin{algorithmic}[1]
   \STATE {\bfseries Input:} Task $\mathcal{D}$, LLM $\mathcal{M}_\theta$, Evaluator $\mathcal{E}$, Ancestor $c_0$
   \STATE \textcolor{gray}{{\bfseries Input:} Env $\mathcal{E}_{env}$, Policy $\pi_\theta$, Reward $\mathcal{R}$, Initial State $\mathbf{s}_0$}
   
   \STATE {\bfseries Initialize:} Evolve Buffer $\mathcal{B}_{e} \leftarrow \{(c_0, s_0, z_0)\}$
   \STATE \textcolor{gray}{{\bfseries Initialize:} Replay Buffer $\mathcal{B}_{r} \leftarrow \emptyset$}
   
   \FOR{$t = 1$ to $T$}
       \STATE \textbf{\textit{// Phase 1: Semantic State Selection}}
       \STATE $(c^{p}_t, z^{p}_t,s^p_t) \leftarrow \text{SelectParent}(\mathcal{B}_{evolve})$
       \STATE \textcolor{gray}{$\mathbf{s}_t \leftarrow \text{Observe}(\mathcal{E}_{env})$;\quad $e_t = \text{Encoder}(\mathbf{s}_t)$}
       
       \STATE \textbf{\textit{// Phase 2: Optimization Direction}}
       \STATE $\tau \leftarrow \text{Rollout}(\mathcal{B}_{e})$;\quad $g_t \leftarrow \text{GradientAgent}(\tau)$
       \STATE \textcolor{gray}{$\tau' \sim \mathcal{B}_{r}$;\quad $\nabla J \leftarrow \text{Estimator}(\tau')$}
       
       \STATE \textbf{\textit{// Phase 3: Experience Distribution}}
       \STATE $E_{ctx} \leftarrow \text{SamplerAgent}(\mathcal{B}_{e}, z^{p}_t, g_t)$
       \STATE \textcolor{gray}{$\tau' \leftarrow \text{PrioritizedSample}(\mathcal{B}_{r})$}
       
       \STATE \textbf{\textit{// Phase 4: Context Construction}}
       \STATE $\Phi_t \leftarrow \text{Compose}(z^{p}_t, g_t, E_{ctx})$
       \STATE \textcolor{gray}{$\Delta \theta \leftarrow \nabla_\theta \mathbb{E}_{\tau} [J(\tau')]$}
       
       \STATE \textbf{\textit{// Phase 5: Code Generation}}
       \STATE $c_t^c \sim \mathcal{M}_\theta(\cdot \mid \Phi_t)$
       \STATE \textcolor{gray}{$\theta' \leftarrow \theta + \alpha\Delta \theta$;\quad $a_t \sim \pi_{\theta'}(\cdot \mid \mathbf{s}_t)$}
       
       \STATE \textbf{\textit{// Phase 6: Evaluation}}
       \STATE $s_t^c \leftarrow \mathcal{E}(c_t^c)$ 
       \STATE \textcolor{gray}{$\mathbf{r}_t \leftarrow \mathcal{R}(\mathbf{s}_t, \mathbf{a}_t)$}
       
       \STATE \textbf{\textit{// Phase 7: Semantic State}}
       \STATE $z_t^c \leftarrow \text{SummaryAgent}(z^{p}_t, c^{c}_t)$
       \STATE \textcolor{gray}{$\mathbf{s}_{t+1} \leftarrow \mathcal{E}_{env}(\mathbf{s}_t,\mathbf{a}_t);\quad e_{t+1}=\text{Encoder}(\mathbf{s}_{t+1})$}
       
       \STATE \textbf{\textit{// Phase 8: Buffer Updat}}
       \STATE $\mathcal{B}_{e} \leftarrow \mathcal{B}_{e} \cup \{(c^{p}_t,s^{p}_t,z^{p}_t;c^c_t, s^c_t, z^c_t)\}$
       \STATE \textcolor{gray}{$\mathcal{B}_{r} \leftarrow \mathcal{B}_{r} \cup \{(\mathbf{s}_t, \mathbf{a}_t, \mathbf{r}_t, \mathbf{s}_{t+1})\}$}
   \ENDFOR
   \STATE {\bfseries Output:} Best solution $c^*$ with the highest score $s^*$
\end{algorithmic}
\end{algorithm}

\subsection{Specialized Context Compression Agents}
While training-free inference-time optimization demonstrates potential in ADRS, current approaches still face significant challenges regarding search efficiency and information density within limited contexts.
Frameworks like OpenEvolve retain the original optimization history within a restricted window, which results in the underutilization of context, leading to search efficiency and suboptimal outcomes.
This challenge underscores the critical need for effective context management in the face of accumulating information.
To address this, we decompose the optimization context into three largely orthogonal semantic dimensions: \textit{semantic state}, \textit{optimization direction}, and \textit{experience distribution}, and employ three specialized agents to distinctively maintain them.

\subsubsection*{\textbf{Summarizer Agent: Semantic State Condensation}}
The Summarizer Agent maintains and condenses the \textit{semantic state}.
It encodes high-dimensional code into a concise, high-level textual abstract that strips away redundancy while preserving critical property.
To ensure the abstract captures both the innovative designs of the offspring and the vital functional segments inherited from the parent, we provide the agent with the parent's abstract $z_{p}$ and the offspring's raw code $c_{c}$.
The Summarizer Agent is tasked with summarizing both novel aspects and preserved elements, formulated as
\vspace{-4px}
\begin{equation}
    z^c \sim \mathcal{M}_{\theta_1}(\cdot \mid \text{Prompt}_{\text{summary}}(z^{p}, c^{c})).
\end{equation}
By transforming raw code differences and similarities into dense semantic descriptions, the Summarizer Agent packs more insights into the limited context window, ensuring that critical historical experience remains accessible throughout entire evolutionary process.

\subsubsection*{\textbf{Navigator Agent: Optimization Direction Distillation}}
The Navigator Agent governs the \textit{optimization direction} by distilling high-level guidance from historical successes and failures.
To ensure the offspring benefits from high-quality past improvements, the agent samples trajectories based on metric variations $\Delta s = s_{parent} - s_{child}$.
These trajectories are weighted and sampled based on $\Delta s$ across three distinct categories: consistent improvement, mixed fluctuation, and consistent decline.
By recording the evolution of abstracts and the corresponding metric shifts, the agent distills directional guidance,
\vspace{-2px}
\begin{align}
    g_t = &\text{GradientAgent}(\{ \tau \sim p(\Delta s)^m_{j=1}\}) \notag \\
    &\sim \mathcal{M}_{\theta_2}(\cdot \mid \text{Prompt}_{\text{summary}}\{(z^p_i, z^c_i, \Delta s_i)\}_{i \in \tau}),
\end{align}
where $m$ is the number of sampled trajectories and $p(\Delta s)$ weights the categories.

By analyzing the correlation between algorithmic changes and metric fluctuations, the Navigator Agent prevents the evolution from repeated futile attempts and steers it toward unexplored high-potential regions, significantly accelerating convergence toward optimal solutions.

\subsubsection*{\textbf{Sampler Agent: Experience Distribution Modulation}}
The Sampler Agent regulates the \textit{experience distribution} by curating a small set of the most informative solutions to serve as quality few-shot exemplars.
To balance exploration and exploitation, the agent selects individuals from the population based on their relevance, diversity, and proven outcomes, considering the parent state $z^{p}$ and the current guidance $g$,
\vspace{-2px}
\begin{equation}
    E_{ctx} \sim \mathcal{M}_{\theta_3}(\cdot \mid \text{Prompt}_{\text{sample}}(\mathcal{B}_{e}, z^{p}, g)).
\end{equation}

By populating the context window with high-value demonstrations rather than random history, the Sampler Agent provides the generator with the most relevant references for the current optimization step while maximizing the utility of each token.

In conclusion, these specialized agents collaboratively transform context management from mere log buildup into an active compression process.
This capability enables the system to conduct deep, long-horizon searches within a fixed-length context window and achieve robust optimization performance with minimal token consumption.

\subsection{Functional Isomorphism with RL}
The efficiency of ContextEvolve comes from its active context compression.
However, beyond this information-theoretic perspective, we observe a profound functional isomorphism between our multi-agent framework and the fundamental components of RL.
Notably, this alignment emerges naturally from the logical decomposition of context management rather than guiding its initial creation. 
Given that RL frameworks are renowned for their high sample efficiency in complex decision spaces, this isomorphism provides strong support for the superior search capabilities and token efficiency of our parameter-blind framework.
Specifically, just as RL agents maximize cumulative rewards by iteratively updating representations, directions, and experiences, our agents collaborate to refine solutions through analogous mechanisms.

\subsubsection*{\textbf{Semantic State as State Representation}}
In RL, raw observations are often high-dimensional and noisy.
Effective learning relies on an encoder that condenses these observations into a compact latent feature vector, capturing the essential dynamics required for downstream decision-making.
Similarly, the Summarizer Agent condenses high-dimensional code artifacts $c$ into a concise semantic abstract $z$,
\begin{equation}
    \text{SummaryAgent}(z^p, c^c) \Leftrightarrow \text{Encoder}(\mathbf{o}_t).
\end{equation}

\subsubsection*{\textbf{Textual Guidance as Policy Gradient}}
Policy gradients in RL derive directional updates from sampled trajectories to maximize the expected return.
In our setting, since the model parameters $\theta$ are frozen, the context prompt $\Phi$ serves as the effective adjustable parameter set. 
The Navigator Agent performs an operation analogous to gradient estimation by distilling improvement directions from weighted historical trajectories,
\begin{equation}
       \text{GradientAgent}(\tau) \Leftrightarrow \nabla_\theta \mathbb{E}_{\tau} [J(\tau)].
\end{equation}

This isomorphism reveals that ContextEvolve performs gradient ascent in the semantic space, offering a directed search mechanism more efficient than random mutation.

\subsubsection*{\textbf{Context Sampling as Prioritized Experience}}
Off-policy RL gains massive efficiency by breaking temporal correlations and reusing past transitions via a Replay Buffer.
The Sampler Agent implements a semantic version of this, retrieving exemplars $  E_{\text{ctx}}  $ conditioned on the current semantic state and directional guidance:
\begin{equation}
    \text{SamplerAgent}(\mathcal{B}_{e}, z^p, g)  \Leftrightarrow \text{PS}(\mathcal{B}_{r}),
\end{equation}

It enables the model learns from the most instructive historical failures and successes rather than the immediate past.

\subsubsection*{\textbf{System-Level Structural Alignment}}
Beyond the context agents, the structure of ContextEvolve mirrors the fundamental architecture of an RL system:
\begin{itemize}[leftmargin=*]
\item \textbf{Generator Agent as Policy Network:}
The generator $\mathcal{M}_\theta$, conditioned on the dynamic context $\Phi$, functions as the stochastic policy $\pi(\cdot | \mathbf{s})$. 
By integrating the compressed state, direction, and exemplars, it executes the generative action that maps the current context to the solution space.
\item \textbf{Evolve Buffer as Replay Buffer:}
The storage of tuples $(c, s, z)$ in $\mathcal{B}_{evolve}$ is functionally equivalent to the replay buffer $\mathcal{D} = \{(\mathbf{s}, a, r, \mathbf{s}')\}$.
This decoupling of data generation from data utilization allows our agents to perform off-policy optimization, extracting global insights from the entire exploration history.
\end{itemize}

In conclusion, this isomorphism reveals that ContextEvolve is not merely a heuristic search method but a principled approximation of an RL system operating in a textual latent space.
By reconstructing the mechanisms of state representation, gradient guidance, and experience replay using natural language agents, ContextEvolve inherits RL-like sample efficiency in a completely training-free setting.

\section{Experiments}
\begin{table*}[t] 
\centering
\caption{Overall performance comparison on ADRS benchmark. We report the key domain-specific metrics along with combined score.}
\vspace{-6px}
\label{tab::overall}
\setlength{\tabcolsep}{2.5pt}
\footnotesize
\begin{tabular}{c|ccc|ccc|ccc|ccc|ccc} 
\toprule 
\multirow{2}{*}{\textbf{Method}} 
& \multicolumn{3}{c|}{\textbf{TS} (100 iters)} 
& \multicolumn{3}{c|}{\textbf{SQL} (100 iters)} 
& \multicolumn{3}{c|}{\textbf{LB} (300 iters)} 
& \multicolumn{3}{c|}{\textbf{SAK} (100 iters)} 
& \multicolumn{3}{c}{\textbf{MP} (100 iters)} \\
& {Make.$\downarrow$} & {Corr.$\uparrow$} & {Score$\uparrow$} 
& {Hit.$\uparrow$} & {Lat.$\downarrow$} & {Score$\uparrow$} 
& {Bal.$\uparrow$} & {Spe.$\uparrow$} & {Score$\uparrow$} 
& {Dens.$\downarrow$} & {Err.$\downarrow$} & {Score$\downarrow$} 
& {Press.$\uparrow$} & {Succ.$\uparrow$} & {Score$\uparrow$} \\
\midrule
\small
Heuristics 
& 38.50 & 1.00 & 25.91 
& 0.56 & 51.86 & 0.53
& 0.25 & 0.20 & 0.13 
& 0.731 & 0.481 & 0.606 
& 20.89 & 1.00 & 21.89 \\

Human-SOTA 
& 32.40 & 1.00 & 30.80 
& 0.69 & 17.43 & 0.66
& 0.24 & 0.43 & 0.14 
& 0.717 & \underline{0.469} & 0.593
& 21.34 & 1.00 & 22.34 \\

LLM One-shot 
& 35.80 & 1.00 & 27.86 
& 0.64 & 0.69 & 0.66 
& 0.25 & 0.20 & 0.13 
& 0.730 & 0.471 & 0.600 
& 21.03 & 0.96 & 21.99 \\

\midrule

GEPA 
& \underline{29.00} & 1.00 & \underline{34.36} 
& \underline{0.72} & \underline{0.65} & \underline{0.73} 
& 0.25 & 0.20 & 0.14
& \textbf{0.627} & 0.575 & 0.602 
& 21.49 & 1.00 & 22.49 \\

OpenEvolve 
& 29.70 & 1.00 & 33.56 
& 0.71 & 2.01 & 0.72
& 0.25 & \underline{0.45} & \underline{0.15} 
& 0.727 & \textbf{0.454} & \underline{0.591}
& \underline{21.67} & 1.00 & \underline{22.67} \\

\textbf{ContextEvolve} 
& \textbf{27.60} & 1.00 & \textbf{36.10} 
& \textbf{0.78} & \textbf{0.56} & \textbf{0.79} 
& \textbf{0.34} & \textbf{0.65} & \textbf{0.20} 
& \underline{0.676} & 0.496 & \textbf{0.586} 
& \textbf{23.02} & 1.00 & \textbf{24.02} \\

\midrule
\textbf{Impr.\%} 
& -4.8 & +0.0 & +5.1
& +8.3 & -13.9 & +8.2
& +36.0 & +44.4 & +33.3
& -7.8 & -9.2 & +0.9
& +6.2 & +0.0 & +6.0 \\

\bottomrule
\end{tabular}
\vspace{-6px}
\end{table*}

\begin{figure*}[t]
    \centering
    \includegraphics[width=0.99\linewidth]{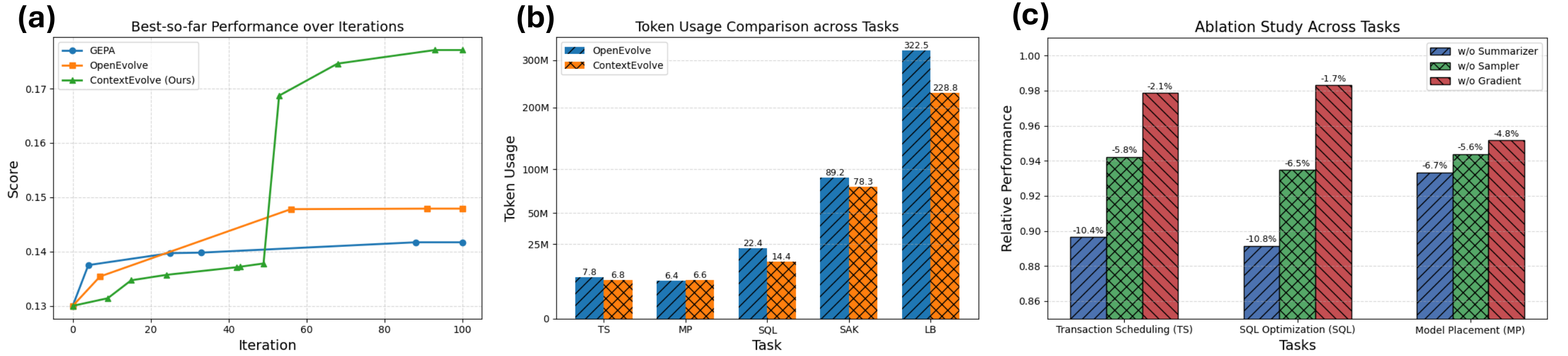}
    \vspace{-6px}
    \caption{
            Efficiency analysis and ablation studies of ContextEvolve.
            \textbf{(a)} Best-so-far performance trajectories over evolutionary iterations in the LB task.
            \textbf{(b)} Cumulative token usage across five ADRS tasks.
            \textbf{(c)} Relative performance of ablated variants.
    }
    \vspace{-8px}
    \label{fig:efficiency}
\end{figure*}

\subsection{Experimental Setup}
We evaluate ContextEvolve on five challenging scenarios from the ADRS benchmark: {Transaction Scheduling (TS)}, SQL Optimization (SQL), {Load Balancing (LB)}, {Sparse Attention Kernel (SAK)}, and {Model Placement (MP)}.
Specific details about these tasks are provided in Appendix~\ref{app:adrs}.

We first compare ContextEvolve against static generation methods, including: (1) \textbf{Heuristics}~\citep{cheng2025barbarians}, representing traditional rule-based algorithms widely deployed in production systems; (2) \textbf{Human-SOTA}~\citep{cheng2024towards,liu2025supporting,yu2025prism,desai2025vattention}, denoting state-of-the-art solutions manually crafted by domain experts; and (3) \textbf{LLM One-shot}, where the model produces a solution in a single pass based on the problem description, serving as the performance lower bound for LLM-based capabilities.
We also compare against advanced LLM-based evolutionary approaches, including (4) \textbf{OpenEvolve} (an open-source implementation of AlphaEvolve)~\cite{openevolve}, and (5) \textbf{GEPA}~\cite{agrawal2025gepa}, a prompt optimization framework that employs reflective mutation and Pareto-based selection.
Since these tasks universally require optimizing trade-offs, we report two competing metrics as well as a weighted combined score which calculated based on the domain-specific importance.
We utilize Qwen3~\citep{yang2025qwen3} as the underlying foundation model for all LLM-based components, including the context agents in our framework and the baseline generators.

\subsection{Overall Performance}
The comparison results on the ADRS benchmark between ContextEvolve and baselines are presented in Table~\ref{tab::overall} where we can derive the following observations:

\begin{itemize}[leftmargin=*]
    \item \textbf{Existing baselines yield sub-optimal solutions due to search inefficiencies.}
    Static generation methods, including Heuristics, Human-SOTA, and LLM One-shot, generally establish the lower bound for performance.
    Especially in the TS task, the Human-SOTA solution lags behind the evolutionary method by over 17\%.
    These methods rely on pre-defined rules or single-pass inference and lack the feedback loops necessary to navigate high-dimensional search spaces. 
    Though evolutinary methods like GEPA and OpenEvolve improve upon static baselines by leveraging iterative feedback, they suffer from inefficient context utilization, limiting the depth of exploration within a fixed budget.
    On average, they always obtain sub-optimal solution and trail ContextEvolve by 6.5\% in overall scores.

    \item \textbf{ContextEvolve displays significant advantages through multi-agent context compression.}
    Our approach leverages multi-agent collaboration and orthogonal context decomposition to actively compress optimization histories, addressing the challenges of unguided search and information bloat.
    Through specialized agents for context compression, it comprehensively surpasses existing baselines across all five ADRS tasks, achieving an average score improvement of 6.5\% over the top baseline.
    Particularly in the LB task, where baselines struggle to optimize the balance metric and only yield minor gains in speed, our method elevates balance by over 36\% while delivering the fastest speed.
    These results underscore the necessity and effectiveness of context compression in evolutionary optimization, enabling outstanding search efficiency with superior token utility.
\end{itemize}

We further analyze the efficiency of our framework by examining both the evolutionary trajectory and the token cost associated with the search process.
As illustrated in Figure~\ref{fig:efficiency}(a), evolutionary baselines converge rapidly within the first 60 iterations and expend over 40\% of total attempts on failed or repetitive exploration to achieve less than a 0.1\% score improvement, indicating a low marginal utility of search.
In contrast, ContextEvolve maintains a continuous upward trajectory by context compression, which updates the best-so-far solution 83.3\% more frequently than baselines.
Notably, when baselines stagnate in local optima, ContextEvolve achieves a 22.4\% score breakthrough and sustain steady improvements.

To validate our context compression strategy, Figure~\ref{fig:efficiency}(b) compares cumulative token usage against OpenEvolve.
A counter-intuitive finding is that while our multi-agent framework requires over $3\times$ more API calls, total token consumption is universally lower, averaging a 17.3\% reduction.
Whereas OpenEvolve concatenates multiple raw codes directly into prompts, ContextEvolve curates fewer high-value exemplars based on condensed semantic states, significantly boosting information density per token.
Furthermore, this efficiency gain scales with task complexity.
For lightweight tasks like Transaction Scheduling (TS) and Model Placement (MP) with short code solutions, the consumption remains comparable (difference $<5\%$).
But for complex tasks like Load Balancing (LB) that involve lengthy implementations, our method saves nearly 30\%, proving its strong handling of high-context loads.

\subsection{Ablation Study}
We conduct an ablation study to validate the efficacy of our multi-agent architecture.
We evaluate variants of ContextEvolve by independently removing each context management agent and illustrate the results in Figure~\ref{fig:efficiency}(c).
The removal of any agent leads to a decline in final solution performance, with average impacts ranked as Summarizer Agent (-9.3\%), Sampler Agent (-6.0\%), and Navigator Agent (-2.9\%).
These results confirm that our architecture effectively manages distinct context dimensions, and the absence of any agent results in a loss of critical context information, thereby reducing search efficiency and overall performance degradation.
Notably, removing the Summarizer Agent yields the most severe regression, pushing performance close to the LLM one-shot baseline. 
Specifically, the scores on TS, SQL, and MP deteriorate by 10.4\%, 10.8\%, and 6.7\%, respectively.
Without such semantic condensation, the limited context window becomes dominated by verbose syntax and low-level details, sharply reducing information density and directly impairs the reasoning capabilities of the Generator Agent.
Furthermore, the absence of concise summaries hinders the Gradient and Sampler Agents from distilling valid optimization directions and curating relevant exemplars, which indirectly destabilizes the overall evolutionary performance through cascading effects in agent cooperation.

\subsection{Case Study: Load Balancing Task}\label{sec:case}

\begin{figure}
    \centering
    \includegraphics[width=0.95\linewidth]{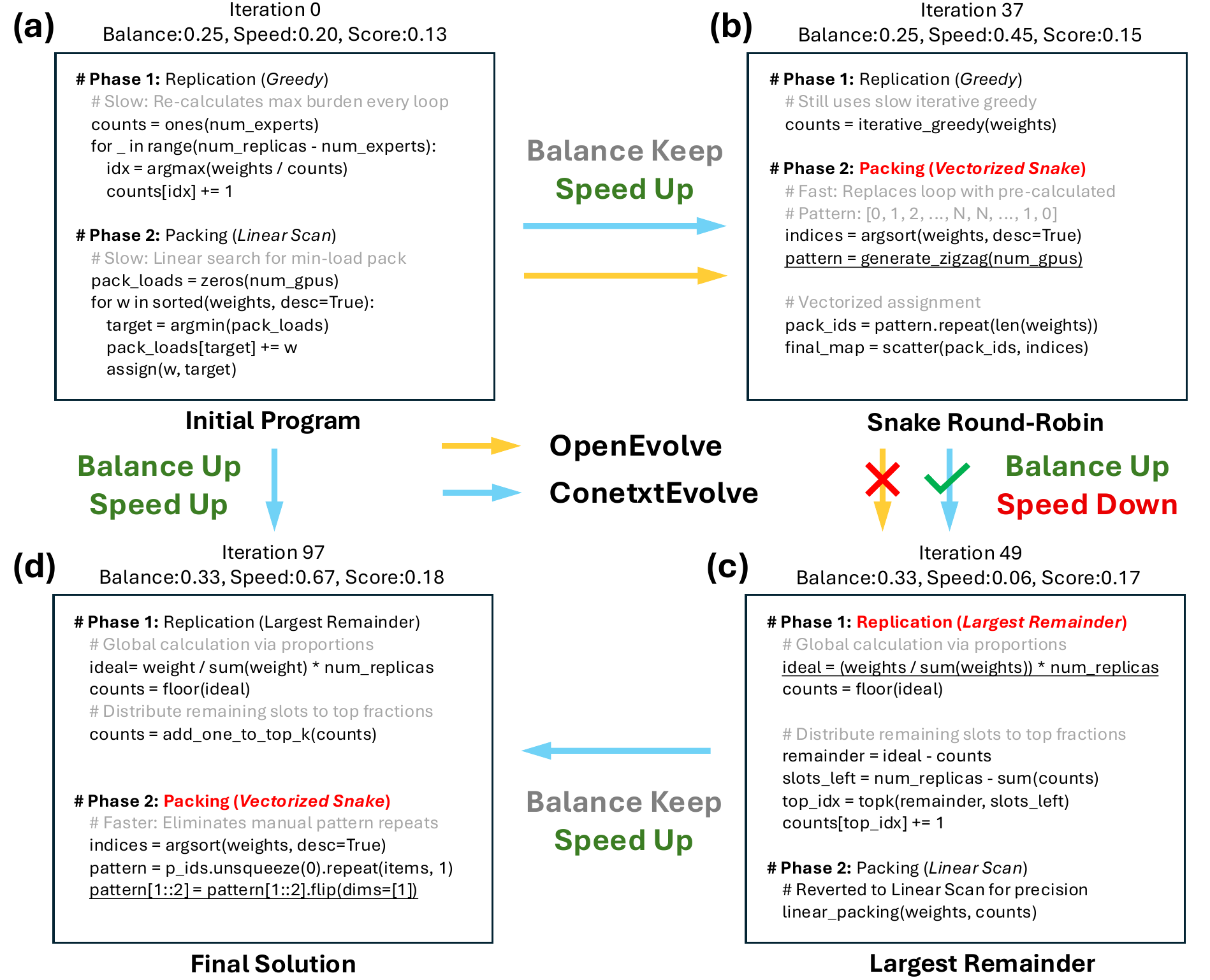}
    \caption{
    \textbf{(a)} The initial solution adopts greedy replication and linear-scan packing, achieving moderate balance but limited speed.
    \textbf{(b)} A vectorized snake round-robin assignment improves runtime while preserving balance.
    \textbf{(c)} Largest-remainder proportional apportionment yields balance gains at the cost of reduced speed.
    \textbf{(d)} The final solution recovers speed without sacrificing balance.
    }
    \label{fig:case}
    \vspace{-10px}
\end{figure}

\begin{figure*}[t]
    \centering
    \includegraphics[width=0.99\linewidth]{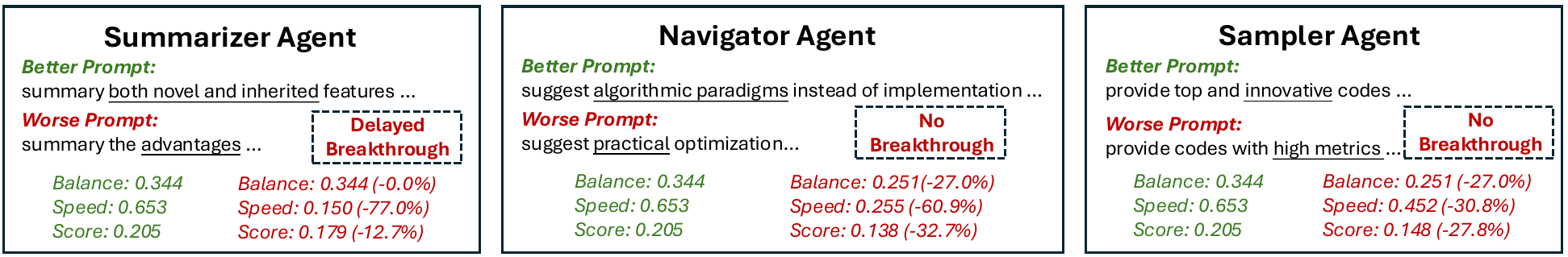}
    \vspace{-6px}
    \caption{Takeaways from prompt perturbations.
    \textbf{(a)} Preserving ancestral traits is as critical as capturing innovation.
    \textbf{(b)} Directional ambiguity guidance outperforms implementation specificity.
    \textbf{(c)} Sampling should prioritize informative semantics, not only high scores.
    }
    \label{fig:takeaway}
    \vspace{-8px}
\end{figure*}

We conduct a case study on the LB task to dissect how ContextEvolve achieves algorithmic breakthroughs compared to evolutionary baselines.
The evolution begins with a standard greedy implementation derived from DeepSeek~\cite{deepseek2024eplb}.
After 10 evolutionary iterations, both OpenEvolve and ContextEvolve independently discover a "Snake Round-Robin" heuristic strategy.
This approach replaces the inefficient linear search in the packing phase with a pre-calculated zigzag assignment pattern, significantly improving assignment speed.
Notably, this variant is not publicly available~\cite{cheng2025barbarians} and is absent from the base model’s training data~\cite{yang2025qwen3}, highlighting the capability of LLM-based evolution to discover advanced algorithmic concepts.

Afterward, the optimization trajectories diverge.
While OpenEvolve generates only minor syntactic variations of the zigzag pattern until the budget is exhausted, ContextEvolve analyzes historical bottlenecks and identifies that heuristic sorting alone fails to handle extreme tail loads.
As illustrated in Figure~\ref{fig:case}(c), by the 49th iteration ContextEvolve proposes a fundamental algorithmic shift in the replication stage, abandoning the iterative greedy approach in favor of a largest remainder proportional apportionment strategy.
Specifically, it calculates the ideal replica count based on the expert's global load proportion, floors this value to ensure minimum allocation, and prioritizes distributing the remaining capacity to experts with the largest fractional remainders.
This rigorous global allocation boosts the balancedness score from 0.25 to 0.33 (+32\%).
Despite the gain in balance, this sophisticated allocation introduces computational overhead via complex tensor operations, initially causing the speed score to drop.
To address this, ContextEvolve retrieves both the previous high-speed solution and the new high-balance solution as in-context exemplars. 
Leveraging these references, the framework re-integrates the searched "Snake Round-Robin" strategy, and further introduces closed-form tensor transformations (see Figure~\ref{fig:case}(d)) to substantially strengthen the speed advantage.
This optimization restores and exceeds the original speed score while maintaining superior balancedness, outperforming the baselines on both metrics simultaneously.

\subsection{Design Takeaways}
Beyond demonstrating performance superiority, we aim to distill high-level design concepts for context management that may generalize to other LLM-based inference-time search methods.
Unlike the structural ablation study, we maintain the macro-level multi-agent architecture but apply specific micro-level perturbations to the prompt design of each agent.
We visualize these prompt modifications and their performance on the Load Balancing (LB) task in Figure~\ref{fig:takeaway}, from which we derive three critical insights.

\textbf{Preserving ancestral traits is as critical as capturing innovation.}
For the Summarizer Agent, effective semantic condensation must balance the extraction of novel modifications with the retention of inherited features.
When we modify the prompt to summarize only the novel changes of an offspring, the performance drops by more than 12.6\%.
Further analysis of the evolutionary trajectory further shows that while the key heuristic strategy (see Section~\ref{sec:case} (c)) still emerges, its discovery is delayed substantially from the 49th to the 87th iteration.
The underlying cause is an amnesia effect: beneficial inherited characteristics are present in the raw code but omitted in the semantic state.
As a result, other agents that rely on summaries repeatedly misinterpret well-explored directions as underexplored opportunities, leading to redundant trials and degraded search efficiency.

\textbf{Directional ambiguity guidance outperforms implementation specificity.}
For the Navigator Agent, ambiguity in high-level direction is beneficial, while over-specification is detrimental.
We modified the gradient prompt to provide specific, practical implementation steps rather than abstract directions, performance significantly drops to 0.138 ($-32.7\%$), failing to find advanced heuristics.
This collapse arises because highly specific instructions implicitly commit the search to a narrow set of edits before candidate generation.
Consequently, the Generator Agent is forced to instantiate a prescribed plan with limited degrees of freedom, behaving more like a code formatter than an explorer of alternative algorithmic branches.
This prematurely narrows the effective solution space, suppresses generative diversity, and makes the optimization trajectory prone to local optima, thereby hindering the discovery of new paradigms.

\textbf{Sampling should prioritize informative semantics, not only high scores.}
For the Sampler Agent, high-scoring exemplars are not the sole source of valuable information, as low-quality or even failed individuals often contain the seeds of breakthroughs.
When we restrict the sampler to return only the highest-scoring individuals, the final score decreases to 0.148 ($-27.8\%$), only on par with OpenEvolve, and the run fails to identify advanced heuristics.
A retrospective analysis shows that the decisive inspiration for the final breakthrough on balancedness in the original ContextEvolve comes from a failed "vectorized binary-search thresholding strategy," which received a score of 0 due to implementation errors.
Although functionally broken, its underlying logic of heuristic allocation was semantically innovative, motivating the generation of the best-performing "Largest Remainder" strategy.
Through semantic state analysis, the Sampler Agent should carefully extract promising concepts even from low-scoring individuals, allowing the system to iterate on flawed but brilliant ideas and preventing the premature discard of potentially transformative insights due to imperfect preliminary implementations.

\section{Discussion}

In this work, we present ContextEvolve, a multi-agent framework that addresses the inefficiency of training-free evolutionary search for systems code optimization.
By decomposing the optimization history into orthogonal dimensions, our approach effectively overcomes the information bottlenecks.
This architecture establishes a functional isomorphism with reinforcement learning, indicating its high search efficiency within parameter-blind setting.
Empirical results on the ADRS benchmark demonstrate that ContextEvolve significantly outperforms state-of-the-art baselines in solution quality while substantially reducing token consumption.

Despite these advancements, several avenues remain for future investigation.
Our research focuses primarily on functionally isolated algorithm.
Scaling ContextEvolve to optimize large-scale codebases with complex inter-module dependencies requires further study.
Moreover, developing stabilizing mechanisms to mitigate the high variance exacerbated by LLM is critical for ensuring consistent outcomes.
Finally, we will investigate evaluation mechanisms to promote the discovery of diversity algorithmic diverse algorithmic paradigms.

\nocite{langley00}

\section*{Impact Statement}
This paper presents work whose goal is to advance the field of machine learning.
There are many potential societal consequences of our work, none of which we feel must be specifically highlighted here.

\bibliography{icml26/icml26}
\bibliographystyle{icml26/icml2026}

\newpage
\appendix
\onecolumn
\section{ADRS Benchmark}\label{app:adrs}
The ADRS benchmark contains five challenging optimization tasks that demonstrate the versatility and effectiveness of ContextEvolve in different contexts.
Below, each task is described along with its specific objectives and the trade-offs involved in optimizing the solution.

\begin{itemize}
\item Transaction Scheduling (TS):
This task involves optimizing the execution order of transactions in a database system to minimize the total execution time, also known as the makespan (Make.).
The optimization problem requires balancing the trade-off between transaction order and latency.
The goal is to maximize the correctness rate (Corr.) and throughput.

\item SQL Optimization (SQL):
In SQL Optimization, the objective is to reorder rows and fields of a table to maximize the hit rate in a Key-Value (KV) cache, which ultimately improves the speed of SQL query inference.
The challenge lies in balancing the prefix cache hit rate (Hit.) against the latency (Lat.) of the reordering algorithm.

\item Load Balancing (LB):
The Load Balancing task focuses on optimizing the distribution of computational load across multiple GPUs, such as in a Mixture of Experts (MoE) model.
The key objective is to maximize the load imbalance (Bal.), ensuring that each GPU handles a proportionate share of the computational workload while also maximizing the speed (Spe.) of rebalancing the load across GPUs when changes occur.

\item Sparse Attention Kernel (SAK):
This task aims to optimize sparse attention mechanisms in neural network models, where the goal is to strike a balance between the density of the active indices in the attention mask and the relative error introduced by this sparsity.
The optimization involves designing attention masks that minimize the combined loss of density (Dens.) and relative error (Err.).

\item Model Placement (MP):
The Model Placement task involves optimizing the placement of large machine learning models across multiple GPUs to reduce contention and ensure efficient resource utilization.
The optimization focuses on minimizing the KV pressure ratio (KVPR) across GPUs, which is a measure of cache contention.
The task seeks to reduce the KVPR (higher Press.) to improve the performance of model inference while ensuring successful (Succ.) utilization of GPUs.
\end{itemize}

\section{Best Evolved Code}
In this section, we present specific examples of the highest-performing code solutions discovered by ContextEvolve during the evolutionary process.
Specifically, Figure~\ref{fig:code_comparison} provides a side-by-side comparison of the evolutionary breakthrough in the Load Balancing (LB) task.
The initial solution relies on iterative Python loops, while the best solution discovered by ContextEvolve utilizes vectorized tensor operations and a rigorous proportional allocation strategy.
The initial codes and the best evolved code for all the five tasks are provided in our code repository.
These artifacts demonstrate the ability of our framework to discover complex algorithmic logic, such as proportional apportionment and heap-based optimization, which are absent in the initial seed code.

\definecolor{codegreen}{rgb}{0,0.6,0}
\definecolor{codegray}{rgb}{0.5,0.5,0.5}
\definecolor{codepurple}{rgb}{0.58,0,0.82}
\definecolor{backcolour}{rgb}{0.95,0.95,0.92}
\definecolor{highlightbg}{rgb}{0.85, 1.0, 0.85} %

\lstdefinestyle{mystyle}{
    backgroundcolor=\color{backcolour},   
    commentstyle=\color{codegreen},
    keywordstyle=\color{magenta},
    numberstyle=\tiny\color{codegray},
    stringstyle=\color{codepurple},
    basicstyle=\ttfamily\scriptsize, %
    breakatwhitespace=false,         
    breaklines=true,                 
    captionpos=b,                    
    keepspaces=true,                 
    numbers=left,                    
    numbersep=4pt,                  
    showspaces=false,                
    showstringspaces=false,
    showtabs=false,                  
    tabsize=2,
    escapechar=| %
}

\lstset{style=mystyle}

\begin{figure*}[t]
    \centering
    \begin{minipage}[t]{0.48\textwidth}
        \centering
        \textbf{(a) Initial Code} \\
        \textit{\scriptsize{Characteristics: Iterative Greedy, Linear Scan, Python Loops}}
        
\begin{lstlisting}[language=Python]
def balanced_packing(weight, num_packs):
    """
    Greedy Linear Scan Strategy
    Complexity: O(Num_Layers * Num_Groups)
    """
    num_layers, num_groups = weight.shape
    groups_per_pack = num_groups // num_packs
    
    # ... setup code omitted ...

    # SLOW: Python-level nested loops
    for i in range(num_layers):
        pack_weights = [0] * num_packs
        pack_items = [0] * num_packs
        
        # Iterating over every group
        for group in indices[i]:
            # Linear scan to find min-load pack
            pack = min(
                (k for k in range(num_packs) 
                 if pack_items[k] < groups_per_pack),
                key=pack_weights.__getitem__,
            )
            
            # Scalar assignment (High Overhead)
            pack_index[i, group] = pack
            pack_weights[pack] += weight[i, group]
            
    return pack_index, rank_in_pack

def replicate_experts(weight, num_phy):
    """
    Iterative Greedy Replication
    Complexity: O(Num_Replicas * Num_Layers)
    """
    n, num_log = weight.shape
    # ... setup code omitted ...
    
    logcnt = torch.ones(...)
    
    # SLOW: Loop once per extra replica needed
    for i in range(num_log, num_phy):
        # Recalculate max load at every step
        scores = weight / logcnt
        idx = scores.max(dim=-1).indices
        
        # Incremental update
        phy2log[:, i] = idx
        rank[:, i] = logcnt[..., idx]
        logcnt[..., idx] += 1
        
    return phy2log, rank, logcnt
\end{lstlisting}
    \end{minipage}
    \hfill
    \begin{minipage}[t]{0.48\textwidth}
        \centering
        \textbf{(b) Best Evolved Code (ContextEvolve)} \\
        \textit{\scriptsize{Characteristics: Snake Heuristic, Proportional, Vectorized}}
        
\begin{lstlisting}[language=Python]
def balanced_packing(weight, num_packs):
    """
    Vectorized Snake Round-Robin Strategy
    Complexity: O(1) Tensor Operations
    """
    num_layers, num_items = weight.shape
    device = weight.device
    
    # Sort weights descending
    sorted_indices = torch.argsort(weight, descending=True)

    # |\hi{INNOVATION 1: Snake Pattern Generation}|
    # Pattern: 0, 1, 2, ..., N, N, ..., 1, 0
    p_ids = torch.arange(num_packs, device=device)
    pattern = p_ids.unsqueeze(0).repeat(items_per_pack, 1)
    # Flip every second row to create zig-zag
    pattern[1::2] = pattern[1::2].flip(dims=[1])

    # Flatten to match items
    pack_ids = pattern.flatten()[:num_items]
    
    # |\hi{INNOVATION 2: Vectorized Assignment}|
    # Eliminates Python loops entirely
    pack_index.scatter_(1, sorted_indices, pack_ids)
    
    return pack_index, rank_in_pack

def replicate_experts(weight, num_phy):
    """
    Proportional Apportionment (Largest Remainder)
    Complexity: O(1) via Sort/TopK
    """
    num_layers, num_log = weight.shape
    
    # |\hi{INNOVATION 3: Global Calculation}|
    # Calculate ideal share based on weight ratio
    total = weight.sum(dim=1, keepdim=True)
    ideal = (weight * num_phy) / total
    
    # Assign integer parts immediately
    logcnt = ideal.floor().to(torch.int64)

    # |\hi{INNOVATION 4: Deficit Filling}|
    # Distribute remaining slots based on fractions
    deficit = num_phy - logcnt.sum(dim=1, keepdim=True)
    
    # Sort by marginal gain (approximation)
    scores = weight / (logcnt + 1).float()
    _, sorted_indices = torch.sort(scores, desc=True)
    
    # Vectorized allocation of remainder
    rank_idx = torch.arange(num_log).expand(num_layers, -1)
    mask = rank_idx < deficit
    logcnt.scatter_add_(1, sorted_indices, mask.long())

    return build_mapping(logcnt) # Helper omitted
\end{lstlisting}
    \end{minipage}
    \caption{Code evolution in the Load Balancing task. (a) The initial baseline uses inefficient iterative loops for both packing and replication. (b) The best code evolved by ContextEvolve introduces \textbf{Vectorized Snake Round-Robin} (Green Highlights) to maximize speed and \textbf{Proportional Apportionment} (Blue Highlights) to maximize load balance. These algorithmic breakthroughs resulted in a 33.3\% improvement in the combined score.}
    \label{fig:code_comparison}
\end{figure*}

\end{document}